\documentclass[preprint,12pt,authoryear]{elsarticle}

\usepackage{booktabs}
\usepackage{multirow}
\usepackage{caption}
\usepackage{subcaption}
\usepackage{rotating}
\usepackage{booktabs,threeparttable}
\usepackage{makecell}
\usepackage{url}

\usepackage{amssymb}
\usepackage{xcolor}
\usepackage{amsmath}

\usepackage{tikz}
\usepackage{pgfplots}
\usepackage{cleveref}
\pgfplotsset{compat=1.15}
\usetikzlibrary{shapes.geometric, patterns, fit, positioning}
\usepgfplotslibrary{groupplots}

\tikzset{fit margins/.style={/tikz/afit/.cd,#1,
    /tikz/.cd,
    inner xsep=\pgfkeysvalueof{/tikz/afit/left}+\pgfkeysvalueof{/tikz/afit/right},
    inner ysep=\pgfkeysvalueof{/tikz/afit/top}+\pgfkeysvalueof{/tikz/afit/bottom},
    xshift=-\pgfkeysvalueof{/tikz/afit/left}+\pgfkeysvalueof{/tikz/afit/right},
    yshift=-\pgfkeysvalueof{/tikz/afit/bottom}+\pgfkeysvalueof{/tikz/afit/top}},
    afit/.cd,left/.initial=2pt,right/.initial=2pt,bottom/.initial=2pt,top/.initial=2pt}
    
\DeclareRobustCommand{\bbone}{\text{\usefont{U}{bbold}{m}{n}1}}

\usepackage[acronym]{glossaries}

\newacronym{cinn}{cINN}{Conditional Invertible Neural Network}
\newacronym{gefcom}{GEFCom2014}{Global Energy Forecasting Competition 2014}
\newacronym{rmse}{RMSE}{Root Mean Squared Error}
\newacronym{mae}{MAE}{Mean Absolute Error}
\newacronym{crps}{CRPS}{Continuous Ranked Probability Score}
\newacronym{pl}{PL}{Pinball Loss}
\newacronym{pdf}{PDF}{probability density function}
\newacronym{cdf}{CDF}{cumulative distribution function}
\newacronym{nn}{NN}{Feed-Forward Neural Network}
\newacronym{pi}{PI}{Prediction Interval}
\newacronym{rf}{RF}{Random Forest}
\newacronym{lr}{LR}{Linear Regression}
\newacronym{xgb}{XGBoost}{eXtreme Gradient Boosting}
\newacronym{nhits}{N-HITS}{Neural Hierarchical Interpolation for Time Series Forecasting}
\newacronym{tft}{TFT}{Temporal Fusion Transformer}
\newacronym{nnqf}{NNQF}{Nearest Neighbour Quantile Filter}
\newacronym{qrnn}{QRNN}{Quantile Regression Neural Network}

\journal{International Journal of Forecasting}
\makeglossaries
\begin{document}

\begin{frontmatter}



\title{Creating Probabilistic Forecasts from Arbitrary Deterministic Forecasts using Conditional Invertible Neural Networks}


\author[kit]{Kaleb Phipps\corref{cor1}}
\ead{kaleb.phipps@kit.edu}
\author[kit]{Benedikt Heidrich}
\author[kit]{Marian Turowski}
\author[kit]{Moritz Wittig}
\author[kit]{Ralf Mikut}
\author[kit]{Veit Hagenmeyer}

\cortext[cor1]{Corresponding author.}
\affiliation[kit]{organization={Karlsruhe Institute for Technology},
            addressline={Institute for Automation and Applied Informatics}, 
            city={Eggenstein-Leopoldshafen},
            postcode={76344}, 
            country={Germany}}

\begin{abstract}
In various applications, probabilistic forecasts are required to quantify the inherent uncertainty associated with the forecast.
However, numerous modern forecasting methods are still designed to create deterministic forecasts. Transforming these deterministic forecasts into probabilistic forecasts is often challenging and based on numerous assumptions that may not hold in real-world situations. Therefore, the present article proposes a novel approach for creating probabilistic forecasts from arbitrary deterministic forecasts. In order to implement this approach, we use a conditional Invertible Neural Network (cINN). More specifically, we apply a cINN to learn the underlying distribution of the data and then combine the uncertainty from this distribution with an arbitrary deterministic forecast to generate accurate probabilistic forecasts. Our approach enables the simple creation of probabilistic forecasts without complicated statistical loss functions or further assumptions. Besides showing the mathematical validity of our approach, we empirically show that our approach noticeably outperforms traditional methods for including uncertainty in deterministic forecasts and generally outperforms state-of-the-art probabilistic forecasting benchmarks. 
\end{abstract}


\end{frontmatter}
\begin{sloppypar}
\section{Introduction}
\label{sec:introduction}
To quantify the inherent uncertainty associated with any prediction of the future, forecasts must be probabilistic \citep{gneiting2007probabilistic, raftery2016use}. These probabilistic forecasts are crucial for many applications such as stabilising energy systems \citep{dannecker2015}, managing congestion in traffic systems \citep{Liu2021}, or sizing servers of web applications to cope with a certain number of daily visits \citep{koochali2019probabilistic}. Despite this necessity for probabilistic forecasts, many forecasting methods are still deterministic \citep{petropoulos2022forecasting}. Even state-of-the-art forecasting methods such as N-BEATS \citep{oreshkin2019n}, N-HiTS \citep{challu2022n}, and Temporal Fusion Transformers \citep{lim2021temporal}, are still designed only to generate deterministic forecasts by default. Extending these models to create probabilistic forecasts would significantly increase their applicability but is challenging and may involve custom loss functions or novel architectures. 

One solution to overcome these challenges is to create probabilistic forecasts based on these existing deterministic forecasts. For many years such forecasts have been created by assuming a Gaussian distribution based on the mean and variance of the deterministic forecast errors and using this assumption to calculate resulting quantiles \citep{hyndman2018forecasting}. Moreover, such probabilistic forecasts can be created by applying the Bayesian theory of probability to a deterministic method \citep{krzysztofowicz1999bayesian} or considering Monte-Carlo sampling methods \citep{caflisch1998monte}. Whilst these methods have been effective in the past, their successful implementation requires a fundamental understanding of statistics and is based on a series of mathematical assumptions that may not hold in a real-world setting. Ideally, such probabilistic forecasts should be created directly from arbitrary deterministic forecasts, without any further statistical assumptions, and without modifying the existing deterministic base forecaster in any way.

Therefore, in the present article, we present an easy-to-use approach that creates probabilistic forecasts from arbitrary deterministic forecasts by using a \acrfull{cinn} to learn the underlying distribution of the time series data. Since time series have an inherent component of randomness \citep{hyndman2018forecasting}, we propose using this uncertainty within the distribution of the time series data to create probabilistic forecasts. With our approach, we first map the unknown distribution of the underlying time series data to a known and tractable distribution by applying a \acrshort{cinn}. Then, we use the output of a trained arbitrary deterministic forecast as an input to the \acrshort{cinn} and consider the representation of this forecast in the known and tractable distribution. We then analyse the neighbourhood of this representation in the known and tractable distribution to quantify the uncertainty associated with the representation. Finally, we use the backward pass of the \acrshort{cinn} to convert this uncertainty information into the forecast. 

Thus, the main contribution of the present article is threefold. First, we provide a novel approach for creating probabilistic forecasts from arbitrary deterministic forecasts without modifying this deterministic forecast in any way. Second, we mathematically show the validity of the proposed approach. Third, we empirically evaluate the approach using different data sets from various domains. In this empirical evaluation, we consider the improvement over a baseline that assumes a Gaussian distribution of deterministic errors to create probabilistic forecasts, compare our approach to state-of-the-art benchmarks, and recreate the \acrfull{gefcom} competition setting.

The remainder of our article is structured as follows. First, we present related work and highlight the research gap that the present article addresses in \Cref{sec:related_work}. In \Cref{sec:probINN}, we then explain our approach in detail and provide mathematical proof for its validity. Afterwards, we highlight how we use a \acrshort{cinn} to enable probabilistic forecasts from an arbitrary deterministic forecast. We detail the experimental setup in \Cref{sec:experimental_setup}, before presenting our results in \Cref{sec:evaluation}. In \Cref{sec:discussion} we discuss our evaluation and key insights. Finally, we conclude and suggest possible directions for future work in \Cref{sec:conclusion}.

\section{Related Work}
\label{sec:related_work}

Our article is closely related to two research fields, namely previous work that enables probabilistic forecasts from deterministic forecasts and previous work focusing on probabilistic forecasts using a \acrshort{cinn}. In this section, we present and discuss related work from both fields and highlight the research gap addressed by the present article.

\paragraph{Enabling Probabilistic Forecasts} Determining the uncertainty associated with a point prediction is one of the key research areas of uncertainty quantification \citep{smith2013uncertainty}. Therefore, multiple methods exist, including error analysis and the assumption of a Gaussian distribution \citep{hyndman2018forecasting}, a Bayesian method involving assumed priors \citep{krzysztofowicz1999bayesian, kaplan2021bayesian, raftery1997bayesian}, integrating uncertainty into the prediction via an ensemble of predictions \citep{toth2003probability, cramer2021evaluation}, and considering uncertainty through Monte Carlo sampling approaches or similar \citep{caflisch1998monte, camporeale2017adaptive, xiu2010numerical}. Another, more recent method, uses a custom loss function that considers both calibration and sharpness to generate probabilistic forecasts based on deterministic forecasts \citep{camporeale2019generation}. 

Despite their effectiveness, these methods have several limitations. First, they all rely on certain assumptions, either regarding the Gaussian distribution, the prior distribution, or the belief that an ensemble accurately represents the uncertainty in the forecast. Second, most methods are based on statistical methods and, therefore, require a fundamental understanding of statistics to be implemented. Third, these methods, especially the ensemble and Monte Carlo approaches, may be computationally expensive. Furthermore, none of the mentioned methods consider the uncertainty associated with the underlying data distribution when creating probabilistic forecasts.

\paragraph{Probabilistic Forecasts using cINNs}
To create probabilistic forecasts, \acrshortpl{cinn}, also referred to as normalising flows \citep{Ardizzone2019a}, are combined with other machine learning methods. \citet{arpogaus2022short}, for example, apply normalising flows to learn the parameters of Bernstein polynomials, which are in turn used to create a probabilistic forecast. Moreover, \citet{rasul2020multivariate} combine normalising flows with recurrent neural networks to create probabilistic forecasts. Normalising flows are also combined with quantile regression networks and copulas \citep{wen2019deep}, or used to create a conditional approximation of a Gaussian mixture model \citep{jamgochian2022conditional} to improve the accuracy of the resulting probabilistic forecasts. Whilst these methods are all effective, normalising flows are used to enrich existing complex probabilistic forecasting methods, but not to provide probabilistic forecasts themselves.

An alternative method, that directly uses normalising flows in the context of probabilistic forecasts, is to learn multi-dimensional distributions of electricity price differences to predict the trajectory of intraday electricity prices \citep{cramer2022multivariate}. Similarly, normalising flows may be applied multiple times to generate scenario-based probabilistic forecasts \citep{dumas2022deep, zhang2019scenario,ge2020modeling}, or to create a proxy for weather ensemble prediction systems based on numerical weather prediction models \citep{fanfarillo2021probabilistic}. These methods use the generative nature of normalising flows to create multiple predictions drawn from the same distribution. However, the forecasts are only probabilistic as an ensemble and each individual forecast is deterministic. Furthermore, these forecasts rely on the assumption that the underlying learned distribution remains constant, and they do not always take external factors into account. Finally, these methods all focus on directly creating probabilistic forecasts, and not on enabling probabilistic forecasts based on existing arbitrary deterministic forecasts. Therefore, such methods cannot be applied to create probabilistic forecast from existing, well designed, deterministic forecasts.

\paragraph{Research Gap} Altogether, we identify a lack of existing work that directly creates probabilistic forecasts from arbitrary deterministic forecasts, without modifying the existing deterministic forecasting method or making potentially unrealistic statistical assumptions. In the present article, we, therefore, aim to fill this research gap by presenting an easy-to-use approach which we describe in the following section.


\section{Creating Probabilistic Forecasts with a cINN}
\label{sec:probINN}
To create probabilistic forecasts from arbitrary deterministic forecasts, we directly apply the uncertainty in the underlying time series. This uncertainty usually reflects the inherent randomness or unpredictability of the measured underlying system. However, this underlying system typically generates observations of an unknown distribution. Thus, it is challenging to directly include the corresponding uncertainty in a forecast. 

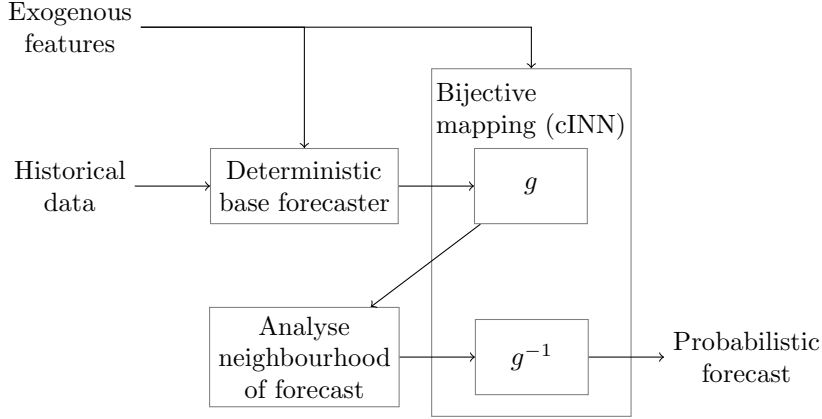
\begin{figure*}[t]
    \footnotesize
    \centering
    \pgfmathdeclarefunction{gauss}{2}{%
	    \pgfmathparse{1/(#2*sqrt(2*pi))*exp(-((x-#1)^2)/(2*#2^2))}%
	    }
	    
    \begin{tikzpicture}[
    	node distance=1.1cm and 1cm, 
    	minimum height=1cm,
    	minimum width=1.5cm,
    	]
         \node (hist) [align=center] {Historical \\ data};
         \node (exog) [align=center, above=of hist] {Exogenous \\ features};

         \node (forecast) [draw=black!50, align=center, right=of hist] {Deterministic \\ base forecaster};
         
         \node (forward) [draw=black!50, align=center, right=of forecast] {$g$};

        \node (sampling) [draw=black!50, align=center, below=of forecast] {Analyse \\ neighbourhood \\ of forecast};

         \node (backward) [draw=black!50, align=center, right=of sampling] {$g^{-1}$};

        \node (probForecast) [align=center, right = of backward] {Probabilistic \\ forecast};

      \node (cINN) [draw=black!50,fit margins={left=8pt,right=8pt,bottom=4pt,top=15pt},  fit=(forward) (backward), label={[anchor=north, align=left, yshift=-2pt, xshift=0pt]Bijective\\mapping (cINN)}] {};

    \draw[->] (hist) -- (forecast);
    \draw[->] (exog) -| (forecast);
    \draw[->] (forecast) -- (forward);
    \draw[->] (exog) -| (cINN);
    \draw[->] (forward) -- (sampling);
    \draw[->] (sampling) -- (backward);
    \draw[->] (backward) -- (probForecast);
    \end{tikzpicture}\vspace{-2mm}
    \caption{Overview of the proposed approach. Exogenous features and historical data are used as inputs for an arbitrary deterministic base forecaster. The resulting deterministic forecast is combined with the exogenous features as inputs to a bijective mapping realised by a \acrshort{cinn}. This mapping creates a representation of the forecast in a known and tractable distribution. We analyse the neighbourhood of this known and tractable representation to gain information about its uncertainty. Finally, we map this representation back to the unknown distribution to create a probabilistic forecast.}
    \label{fig:probINN}
\end{figure*}

To solve this challenge, we aim to find a bijective mapping from the unknown distribution to a known and tractable distribution. This bijective mapping should also consider exogenous features, as shown in \Cref{fig:probINN}. 
With such a mapping $g$, we can map a deterministic forecast from the unknown distribution to its representation in the known and tractable distribution. In the known and tractable distribution, we can analyse the neighbourhood of this representation and can gain information about its uncertainty. Finally, we can map this uncertainty information back to the unknown distribution using the inverse mapping $g^{-1}$ to create probabilistic forecasts. 


In this section, we first mathematically show the validity of our approach that uses the distribution of the underlying data to include uncertainty in a deterministic forecast. We then explain how to apply our approach with a \acrshort{cinn}, starting with the training of this \acrshort{cinn}, before describing how we generate probabilistic forecasts using arbitrary deterministic forecasts.

\subsection{Including Uncertainty from the Underlying Distribution of the Data}
\label{subsec:learning_distribution}

To formally show how a bijective mapping from an unknown distribution in a known and tractable distribution enables probabilistic forecasts using arbitrary deterministic forecasts, we first show that such a mapping exists. Given the existence of this mapping, we show that such a bijective mapping guarantees the equivalence of the uncertainty in the image and the inverse image of the considered mapping.

\paragraph{Bijective Mapping} To introduce the bijective mapping, let us consider a times series $\mathbf{y} = \{y_t\}_{t \in T}$ consisting of $T$ observations as realisations of a random variable $Y \sim f_Y(\mathbf{y})$ with a \acrfull{pdf} $f_Y(\mathbf{y})$ in the realisation space $\mathbb{Y}$. Furthermore, we have a bijective mapping $g : \mathbb{Y} \rightarrow \mathbb{Z}$ from the realisation space $\mathbb{Y}$ to the space of the tractable distribution $\mathbb{Z}$ where $\mathbf{y} \mapsto g(\mathbf{y}, \circ) = \mathbf{z}$, and $g$ being a continuously differentiable function.\footnote{The function $g$ can include further parameters apart from $\mathbf{y}$, such as exogenous information. These further parameters are indicated via $\circ$.} To calculate the \acrshort{pdf} $f_Z(\mathbf{z})$ in terms of $f_Y(\mathbf{y})$, we can apply the change of variables formula \citep{pml2Book}, i.e. \begin{equation}
    f_Z(\mathbf{z}) = f_Y(g^{-1}(\mathbf{z, \circ})) \left| \det\left( \frac{\partial g^{-1}}{\partial \mathbf{z}} \right)  \right|,
    \label{eq:cov}
\end{equation}
where $\frac{\partial g^{-1}}{\partial \mathbf{z}}$ is the Jacobian matrix. Since $g$ is bijective, this equation describes a bijective mapping from the unknown distribution $f_Y(\mathbf{y})$ to the known and tractable distribution $f_Z(\mathbf{z})$. Therefore, the change of variable formula provides us with the required mapping.

\paragraph{Equivalence of Uncertainty} After introducing the bijective mapping, we need to show the equivalence of the uncertainty in the unknown distribution and known tractable distribution when applying \Cref{eq:cov}. More specifically, we show the equivalence of quantiles in both the realisation space and the tractable distribution space, since quantiles serve as a non-parametric representation of the uncertainty. 

To show this equivalence, we first consider the \acrfull{cdf} of the random variable $Z=g(Y,\circ) \sim f_Z(\mathbf{z})$, defined as \begin{equation}
    F_Z(\mathbf{z}) = \int_{-\infty}^\mathbf{z} f_Z(\mathbf{u}) d\mathbf{u}.
    \label{eq:cdf}
\end{equation}
If we use the expression for $f_Z(\mathbf{z})$ from the change of variables formula (\Cref{eq:cov}) in the definition of the \acrshort{cdf} (\Cref{eq:cdf}), we obtain \begin{equation}
    F_Z(\mathbf{z}) = \int_{-\infty}^\mathbf{z} f_Y(g^{-1}(\mathbf{u}, \circ)) \left| \det\left( \frac{\partial g^{-1}}{\partial \mathbf{u}} \right) \right| d\mathbf{u},
    \label{eq:new_cdf}
\end{equation} 
describing the \acrshort{cdf} of $F_Z(\mathbf{z})$ in terms of the \acrshort{cdf} $F_Y(\mathbf{y})$. Since $g$ is per definition a continuously differentiable function, we can apply integration by substitution for multiple variables to rewrite \Cref{eq:new_cdf} as \begin{equation}
    F_Z(\mathbf{z}) = \int_{-\infty}^{g^{-1}(\mathbf{z})} f_Y(\mathbf{v}) d\mathbf{v} = F_Y(g^{-1}(\mathbf{z}, \circ)),
    \label{eq:final_result}
\end{equation}
which is simply the \acrshort{cdf} of $Y$ evaluated at the inverse of $g$. Further, the quantiles $\mathbf{z}_\alpha$ of $Z$ are defined by the inverse of the \acrshort{cdf}, i.e. \begin{align*}
    \mathbf{z}_\alpha = F_Z^{-1}(\alpha) &= \inf { \mathbf{z} \mid F_Z(\mathbf{z}) \geq \alpha } \\
     &= \inf { \mathbf{z} \mid F_Y(g^{-1}(\mathbf{z}, \circ)) \geq \alpha }
\end{align*}
where $\inf$ refers to the infimum, the smallest value of $\mathbf{z}$ that fulfils the condition, and $\alpha \in [0,1]$ is the considered quantile. 
Consequently, if we know that the $\alpha$ quantile of $F_Z$ is $\mathbf{z}_\alpha$, then we can also calculate the $\alpha$ quantile of $F_Y$ as $g^{-1}(\mathbf{z}_\alpha, \circ) = \mathbf{y}_\alpha$. From this follows an equivalence between the quantiles of $Z$ and the quantiles of $Y$, which implies an equivalence in the uncertainty.

Given the mathematical equivalence of the uncertainty in the two considered distributions, we can include uncertainty in a tractable and known distribution $f_Z(\mathbf{z})$ and use the inverse mapping $g^{-1} : \mathbb{Z} \rightarrow \mathbb{Y}$ to map this uncertainty to the original distribution $f_Y(\mathbf{y})$.  To realise this bijective mapping $g$, we use a \acrshort{cinn} \citep{Ardizzone2019a, Heidrich2022}. A \acrshort{cinn} is a neural network that consists of multiple specially designed conditional affine coupling blocks \citep{Ardizzone2019a}. As shown by \citet{Ardizzone2019a}, these coupling blocks ensure that the mapping $g: \mathbb{Y} \rightarrow \mathbb{Z}$ learnt by the \acrshort{cinn} is bijective. Furthermore, with the conditional information the \acrshort{cinn} is able to consider additional information, such as exogenous features or calendar information, when learning the mapping \citep{Ardizzone2019a, Heidrich2022}. As a result, the \acrshort{cinn} learns an approximation of $f_Z(\mathbf{z})$ and a mapping $g$, which is per definition bijective, thus ensuring we can apply \Cref{eq:cov} as described previously. 

\subsection{Applying our Approach}
\label{subsec:method_application}
In the following, we describe how we realise the inclusion of uncertainty with a \acrshort{cinn}.\footnote{The implementation will be made available after the acceptance of the paper.} We first detail how we train a \acrshort{cinn} that learns the distribution of the underlying data. Second, we describe how we use this trained \acrshort{cinn} to generate probabilistic forecasts.

\paragraph{Training} We use a \acrshort{cinn} to realise the continuous differentiable function $g$ described above. In addition to the original realisation $\mathbf{y}$, we also consider conditional information $\mathbf{c}$ as an input to the function $g$. This conditional information always includes calendar features such as time of the day, and day of the week, but depending on the time series may also include additional exogenous features that are available for the forecast period. The aim of the training is to ensure that the \acrshort{cinn} learns the function $g$, so that resulting realisations $\mathbf{z}=g(\mathbf{y},\mathbf{c})$ follow a known and tractable latent space distribution $f_Z(\mathbf{z})$. Therefore, we apply the change of variables formula in the maximum likelihood loss function
\begin{equation} \label{eq:loss}
\mathcal{L} = \frac{\parallel g(\mathbf{y}; \mathbf{c}, \theta)\parallel^2_2 }{2} - \log\mid J \mid,
\end{equation}
where $J = \det(\partial g / \partial \mathbf{y})$ is the determinant of the Jacobian, and $\theta$ is the set of all trainable parameters \citep{Ardizzone2019a, Heidrich2022}. Training a \acrshort{cinn} with this loss function results in a network with the maximum likelihood parameters $\hat{\theta}_{\text{ML}}$ and ensures that the realised latent space distribution $f_Z(\mathbf{z})$ follows a Gaussian distribution \citep{Ardizzone2019a}.

Note that to apply our approach, a deterministic base forecaster must also be trained. However, since our approach simply enables probabilistic forecasts based on arbitrary deterministic forecasts, this deterministic base forecaster can be trained as usual. Thus, we refrain from a more detailed description of the training of the deterministic base forecaster.

\paragraph{Forecasting} To generate probabilistic forecasts, we begin with the output of a deterministic base forecaster $\hat{\mathbf{y}}$. We combine this output with the associated conditional information $\mathbf{c}$ and pass it on through the trained \acrshort{cinn} to obtain a latent space representation of the output, i.e. \begin{equation}
    \hat{\mathbf{z}} = g(\hat{\mathbf{y}},\mathbf{c},\hat{\theta}_{\text{ML}}).
\end{equation}
Given this latent space representation of the deterministic forecast, we explore the uncertainty in the neighbourhood of the forecast with
\begin{equation}\label{eq:explore}
    \tilde{\mathbf{z}}_i = \hat{\mathbf{z}} + \mathbf{r}_i, \quad i=1,\dots, I, \, \, \mathbf{r}_i \sim \mathcal{N}(0, \sigma). 
\end{equation}
Using \Cref{eq:explore}, we select a random noise $\mathbf{r}_i$ from a standard normal distribution with mean $0$ and variance $\sigma$ and add this noise to the realisation $\hat{\mathbf{z}}$. We define the variance used for the sampling process $\sigma$ as the \emph{sampling hyperparameter}, which must be manually optimised using an evaluation metric. Based on the selected $\sigma$, we repeat the sampling process $I$ times to obtain multiple realisations of $\tilde{\mathbf{z}}_i$ that are all similar but not identical to the original forecast in the known latent space distribution. This set of realisations provides a representation of the uncertainty in the neighbourhood of the forecast. Due to the equivalence of uncertainty in both spaces shown in \Cref{subsec:learning_distribution}, we then process each of these samples through the backward pass of the \acrshort{cinn} to include uncertainty around the original representation of the deterministic forecast. Finally, we calculate the quantiles of these samples in the original realisation space. The resulting quantiles represent a probabilistic forecast, derived from the original arbitrary forecast.


\section{Experimental Setup}
\label{sec:experimental_setup}
This section describes the experimental setup we use to evaluate our approach of creating probabilistic forecasts based on arbitrary deterministic forecasts. We first introduce the data used, before explaining the evaluation metrics. Furthermore, we describe the selected base forecasters to generate the deterministic forecasts, introduce the used state-of-the-art probabilistic benchmarks, and detail the implementation of the used \acrshort{cinn}.

\subsection{Data}
\label{subsec:data}
We evaluate our proposed approach on four different data sets. In this section, we briefly introduce each of these data sets before we describe their preprocessing.

The first considered data set is \emph{Electricity}, namely the UCI Electricity Load Dataset\footnote{\url{https://archive.ics.uci.edu/ml/datasets/ElectricityLoadDiagrams20112014}} \citep{Dua.2019}. From this data set, we select the time series \emph{MT\_158} and resample it to an hourly resolution.
The second considered data set, \emph{Bike}, contains hourly records of rented bikes from the UCI Bikesharing Dataset \citep{Fanaee.2013, Dua.2019}.\footnote{\url{https://archive.ics.uci.edu/ml/datasets/bike+sharing+dataset}} 
The third data set, \emph{OPSD}, is a subset of hourly electricity load values for the state of Baden-Württemberg in Germany, taken from the Open Power Systems Data portal.\footnote{\url{https://open-power-system-data.org/}}
The fourth data set, \emph{Price}, is a subset of the data provided for Task~1 in the electricity price track of the \acrfull{gefcom}. The data includes zonal electricity price values recorded for a single location at an hourly resolution \citep{hong2016probabilistic}. 
To enable a thorough comparison of all models, we do not attempt to forecast the single day required for Task~1, but instead, consider a longer period.

We normalise each of the above data sets before creating separate test, validation, and training subsets for the training and testing of our approach. An overview of these splits and the exogenous variables considered for each data set is presented in \Cref{tab:data_overview}. 

\subsection{Evaluation Metrics}
\label{subsec:evaluation metrics}
To evaluate our approach comprehensively, we consider several evaluation metrics, both for deterministic and probabilistic forecasts. In the following, we briefly present these metrics in the order they appear in the evaluation.

\paragraph{Mean Absolute Error} To evaluate the quality of the deterministic base forecasters, we consider the \acrfull{mae}. 
The \acrshort{mae} is given by \begin{equation}
    \operatorname{MAE}(y,\hat{y}) = \frac{1}{n}\sum\limits_{i=1}^n |y_i - \hat{y}_i |
    \label{eq:mae}
\end{equation}
with a true value $y_i$, a forecast value $\hat{y}_i$, and $n$ observations.

\paragraph{Continuous Ranked Probability Score} To evaluate the quality of the probabilistic forecasts, we consider the \acrfull{crps} \citep{matheson1976scoring}. The \acrshort{crps} is a proper scoring rule that measures the calibration and sharpness of a predictive cumulative distribution function $F$ \citep{gneiting2007strictly}. The \acrshort{crps} is defined as
\begin{equation}
\operatorname{CRPS}(F, y) = \int_\mathbb{R} \left( F(z) - \bbone\{y \leq z \} \right) ^2 dz,
\label{eq:crps}
\end{equation}
where $\bbone\{y \leq z\}$ is the indicator function which is one if $y \leq z$ and otherwise zero. Since our approach as well as the benchmarks provide samples drawn from a distribution, we use the sample-based variant of the \acrshort{crps} implemented in the \texttt{properscoring} library.\footnote{https://github.com/properscoring/properscoring}


\paragraph{CRPS Skill Score} To easily determine whether our approach leads to better probabilistic forecasts, we measure the improvement in the \acrshort{crps} compared to a baseline. The resulting CRPS skill score is defined as
\begin{equation}
    \text{CRPS Skill Score} = \left(1 - \frac{\operatorname{CRPS}_{\text{Forecast}}}{\operatorname{CRPS_{\text{Baseline}}}}\right) \cdot 100,
    \label{eq:crps_skill}
\end{equation}
where $\operatorname{CRPS}_{\text{Forecast}}$ is the \acrshort{crps} of the considered forecasting method and $\operatorname{CRPS_{\text{Baseline}}}$ the \acrshort{crps} of a selected baseline.

\paragraph{Normalised Prediction Interval Width} To measure the sharpness of the probabilistic forecasts, we consider the average normalised width of various \acrfullpl{pi}. The normalised average width of the $\beta$-\acrshort{pi} is defined as
\begin{equation}
    |\operatorname{PI}(\beta)| = \frac{1}{\bar{y}} \left( \frac{1}{n} \sum\limits_{i=1}^n |\hat{y}_{i,\frac{1+\beta}{2}} - \hat{y}_{i,\frac{1-\beta}{2}}| \right),
    \label{eq:pi_width}
\end{equation}
where $\hat{y}_{i,\frac{1+\beta}{2}}$ is the predicted upper quantile, $\hat{y}_{i,\frac{1-\beta}{2}}$ the predicted lower quantile for the forecast value $\hat{y}_i$, and $\bar{y}$ the mean of the target time series. We consider the normalised prediction interval width to enable a comparison between different data sets.

\paragraph{Training Time} To evaluate the computational cost, we measure the training time. Thereby, we measure the training time of the \acrshort{cinn} excluding the training time of the deterministic base forecaster. We compare this to the training time of the probabilistic benchmark methods. The training time is measured on a PC with an Intel 3.00 GHz i7-9700 CPU, 32 GB RAM, and no GPU.

\paragraph{Pinball Loss Improvement} In order to evaluate our approach in the recreated \acrshort{gefcom} competition, we consider the scoring mechanism used in this competition. This mechanism relies on the \acrfull{pl}, a scoring rule that minimises the loss when issuing a point forecast for the $\alpha$-quantile \citep{gneiting2022model}. For a set of considered quantiles $Q=[0.01,\dots,0.99]$, the \acrshort{pl} is calculated with  
\begin{equation}\label{eq:pinball}
    \operatorname{PL} = \frac{1}{n\mid Q \mid}\sum_{\alpha \in Q} \sum\limits_{i=1}^n \begin{cases} (y_i - \hat{y}_{i,\alpha}) \cdot \alpha \quad  y_i \geq \hat{y}_{i,\alpha} \\
    (\hat{y}_{i, \alpha}-y_i) \cdot (1-\alpha) \quad \hat{y}_{i, \alpha} > y,        
    \end{cases}
\end{equation}
where $y_i$ is the true value and $\hat{y}_{i,\alpha}$ is the quantile forecast for the quantile $\alpha$. For the \acrshort{gefcom}, the relative improvement of the \acrshort{pl} compared to a given baseline forecast is considered, i.e. \begin{equation}
    \operatorname{PL}_{\%} = \frac{\operatorname{PL}_{\text{Forecast}}}{\operatorname{PL}_{\text{Baseline}}},
    \label{eq:pl_improve}
\end{equation}
where $\operatorname{PL}_{\text{Forecast}}$ is the \acrshort{pl} for the considered forecast and $\operatorname{PL}_{\text{Baseline}}$ the \acrshort{pl} for the baseline provided in the \acrshort{gefcom}.

\subsection{Selected Base Forecasters}
\label{subsec:selected_base}
Our proposed approach can be applied to forecasts from arbitrary deterministic forecasters. Thus, we evaluate our approach on four simple and two state-of-the-art deterministic forecasting methods. 
As simple base deterministic forecasters we consider a \emph{\acrfull{rf}}, a \emph{\acrfull{lr}}, a \emph{\acrfull{nn}}, and the \emph{\acrfull{xgb}} Regressor. The two state-of-the-art deterministic base forecasters are \emph{\acrfull{nhits}} \citep{challu2022n} and \emph{\acrfull{tft}} \citep{lim2021temporal}. We provide implementation details for each of the selected deterministic base forecasters in \Cref{tab:bf_overview}.

As inputs, all base forecasting methods receive $24$ hours of historical information for the target time series and exogenous features for the forecast horizon of $24$ hours. The exogenous features comprise calendar information and, depending on the data set, further exogenous variables, as shown in \Cref{tab:data_overview}.

When applying the base forecasters with the \acrshort{cinn} to create probabilistic forecasts, we manually select the sampling parameter $\sigma$ that minimises the \acrshort{crps} on the validation data set. An overview of the selected sampling parameters is presented in \Cref{tab:sampling_main}. Furthermore, all selected base deterministic forecasters are implemented in a pipeline with pyWATTS \citep{Heidrich2021}\footnote{\url{https://github.com/KIT-IAI/pyWATTS}}.

\subsection{Probabilistic Benchmarks}
\label{subsec:probabilistc_benchmarks}
To assess the quality of the probabilistic forecasts created with our approach, we compare them to three state-of-the-art probabilistic benchmark forecasting methods.

The first considered benchmark method is \emph{DeepAR} \citep{salinas2020deepar}, which is an autoregressive recurrent neural network-based approach for probabilistic forecasting. 
We implement DeepAR using the PyTorch Forecasting library\footnote{\url{https://pytorch-forecasting.readthedocs.io/en/stable/api/}\\\url{pytorch_forecasting.models.deepar.DeepAR.html}}.

The second benchmark method is a \emph{\acrfull{qrnn}}. It trains a \acrshort{nn} to directly forecast a selected quantile, or multiple quantiles, instead of the mean or median \citep{koenker2017handbook}. 
To realise the \acrshort{qrnn}, we use a separate simple feed-forward \acrshort{nn} to forecast each of the selected quantiles training each \acrshort{nn} with the appropriate pinball loss function.
The \acrshort{qrnn} is implemented using TensorFlow\footnote{\url{https://www.tensorflow.org/}} with the Keras\footnote{\url{https://keras.io/}} library and the pinball loss function.

The third benchmark method uses the \emph{\acrfull{nnqf}} proposed by \citet{Ordiano2020}. Similar to the \acrshort{qrnn}, this method also forecasts quantiles. However, instead of using a custom quantile loss function to directly learn the quantiles, the \acrshort{nnqf} finds similar values for each time step based on similarity in the target variable to determine quantiles in the data. A forecasting method is then trained to predict these calculated quantiles \citep{Ordiano2020}.
To realise the \acrshort{nnqf}, we use a multi-layer feed-forward \acrshort{nn} with one output per quantile, which is implemented using sklearn \citep{Pedregosa2011} and pyWATTS \citep{Heidrich2021}.

\subsection{Used cINN}
\label{subsec:network_params}
In the evaluation, we use the same \acrshort{cinn} architecture (see \Cref{tab:cINNHyperparams}) for each of the considered data sets. It is based on GLOW coupling layers that consider conditional input \citep{Kingma2018}. Similar to \citet{Ardizzone2019a,Heidrich2022}, the conditional input is provided by a fully connected \acrshort{nn}, which uses the same exogenous information available to the base forecaster as conditional information (see \Cref{tab:data_overview}). We detail the implementation information for the used \acrshort{cinn} in \ref{appendix:implementation_cinn}. 
We implement the \acrshort{cinn} in a pipeline with pyWATTS \citep{Heidrich2021}.


\section{Evaluation}
\label{sec:evaluation}
Given the experimental setup, we present the evaluation of the proposed approach in this section. We first report the deterministic performance of each base forecaster since they affect the quality of the resulting probabilistic forecasts. We then consider the improvement of the proposed approach over a baseline that assumes a Gaussian error distribution, its performance compared to state-of-the-art probabilistic benchmark forecasts, and how it would have placed in the \acrshort{gefcom}.

\subsection{Deterministic Results}
\label{subsec:results_deterministic}
Since the quality of the deterministic base forecaster influences the resulting probabilistic forecast when combined with a \acrshort{cinn}, we briefly evaluate the deterministic performance of the selected base forecasters. We report the average \acrshort{mae} (\Cref{eq:mae}) over five runs in \Cref{tab:deterministic_mae}.


\begin{table}
    \centering
    \footnotesize
    \caption{The average \acrshort{mae} (\Cref{eq:mae}) on the test data sets across five runs for each of the considered base forecasters introduced in \Cref{subsec:selected_base}. The lowest error for each data set is highlighted in bold.}
    \label{tab:deterministic_mae}
    \begin{tabular}{l|cccccc}
        \toprule
        Data set & RF & LR & NN & XGBoost & N-HITS & TFT \\
                \midrule
        Electricity & 0.2566 & 0.3892 & 0.2535 & \textbf{0.2502} & 0.2724 & 0.2549 \\
        Bike & 0.3076 & 0.4223 & 0.2998 & \textbf{0.2885} & 0.4359 & 0.3834 \\
        OPSD & 0.1549 & 0.2098 & \textbf{0.1272} & 0.1273 & 0.1732 & 0.1514  \\
        Price & 0.2863 & \textbf{0.2585} & 0.2595 & 0.2639 & 0.2983 & 0.3316 \\
        \bottomrule
    \end{tabular}
\end{table}

We see that different deterministic base forecasters perform best depending on the data set considered. However, either the \acrshort{nn} or \acrshort{xgb} achieve the lowest error for three of the four data sets. Furthermore, the size of the error is usually similar for all forecasters in one data set. Finally, the advanced base forecasters such as NHiTS and the TFT, often perform similarly or worse than the four simple base forecasters.

\subsection{Improvement Over Baseline}
\label{subsec:results_improement}
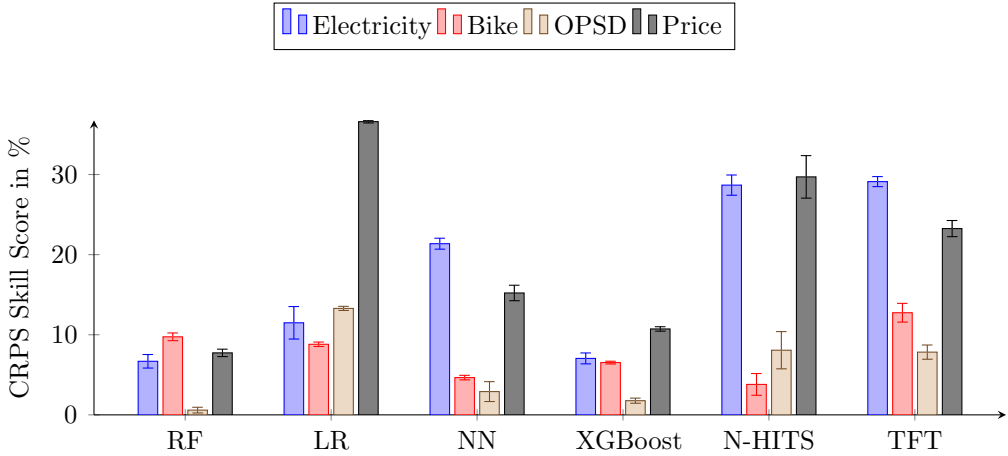
\begin{figure}
    \centering
    \begin{tikzpicture}
    \begin{axis} [
        height = 5.5cm,
        width = 1\textwidth,
        ybar,
        ylabel=CRPS Skill Score in \%,
        ylabel style={font=\footnotesize},
        table/col sep=comma,
        tick label style={font=\footnotesize},
        bar width=0.26cm,
        symbolic x coords ={RF, LR, NN, XGBoost, N-HITS, TFT},
        xticklabel style={text width=1.25cm,align=center, font=\footnotesize},
        xtick=data,
        axis lines=left,
        enlarge x limits = 0.125,
        ymin=0,
        legend style = {
            font=\footnotesize,
            at={(0.45,1.4)},
            anchor=north,
        },
        legend cell align={left},
        legend columns=4
        ]
        \addplot+[error bars/.cd, y dir=both, y explicit]
            table[x=Method,y=Electricity_mean, y error plus=Electricity_std, y error minus=Electricity_std] {results/Skill_Score_Results.csv}; 
        \addlegendentry{Electricity}
            
        \addplot+[error bars/.cd, y dir=both, y explicit]
            table[x=Method,y=Bike_mean, y error plus=Bike_std, y error minus=Bike_std] {results/Skill_Score_Results.csv}; 
        \addlegendentry{Bike}

        \addplot+[error bars/.cd, y dir=both, y explicit]
            table[x=Method,y=OPSD_mean, y error plus=OPSD_std, y error minus=OPSD_std] {results/Skill_Score_Results.csv}; 
        \addlegendentry{OPSD}

        \addplot+[error bars/.cd, y dir=both, y explicit]
            table[x=Method,y=Price_mean, y error plus=Price_std, y error minus=Price_std] {results/Skill_Score_Results.csv}; 
        \addlegendentry{Price}

    
    \end{axis}
\end{tikzpicture}
    \caption{The \acrshort{crps} Skill Score (\Cref{eq:crps_skill}) for all base forecasters introduced in \Cref{subsec:selected_base} and the four considered data sets. The skill score indicates the average improvement on the test data set over five runs compared to the baseline which assumes a Gaussian distribution with the same base forecaster. The error bars indicate the variance across the runs.}
    \label{fig:skill_multi-data}
\end{figure}

To assess the improvement obtained with our approach, we compare it to a baseline. The baseline assumes that the errors from the deterministic forecast follow a Gaussian distribution and generates probabilistic forecasts based on this assumption. We plot the \acrshort{crps} Skill Score (\Cref{eq:crps_skill}) for each base forecaster and every data set in \Cref{fig:skill_multi-data} (see \Cref{tab:crps_raw} for the absolute \acrshort{crps}).

We observe that our approach noticeably improves the performance of all base forecasters across all data sets. The improvement is most noticeable for the electricity and price data sets, where the proposed approach achieves an average improvement of $17.40\%$ and $20.54\%$ respectively. The largest improvement of $38.60\%$ has the \acrshort{lr} base forecaster on the price data set. The improvement is the smallest for the OPSD data set, with our approach only yielding an average improvement of $5.74\%$ and the smallest improvement of $0.59\%$ for the \acrshort{rf} base forecaster. Furthermore, we observe that the improvement is dependent on both the data set and the base forecaster used. The observed improvements in \acrshort{crps} vary for a given base forecaster across the data sets, but also across all base forecasters for each data set.

 
\subsection{Comparison to Benchmarks}
\label{subsection:comparison_direct}
In addition to the comparison to baseline assuming a Gaussian distribution, we also compare our approach to the selected state-of-the-art probabilistic forecasting benchmarks with respect to the quality of the probabilistic forecast, its sharpness, and its training time.

\begin{table}
    \centering
    \footnotesize
    \caption{The \acrshort{crps} (\Cref{eq:crps}) calculated on the test data set and averaged over five runs. The three base forecasters from \Cref{subsec:selected_base} with the best performance using the \acrshort{cinn} in our approach are selected and compared against the three probabilistic benchmark forecasting methods from \Cref{subsec:probabilistc_benchmarks} for each data set. The overall best probabilistic forecast is highlighted in bold.}
    \label{tab:crps_benchmark}
    \begin{threeparttable}
    \begin{tabular}{l|cccccc}
        \toprule
        Data Set & \multicolumn{3}{c}{cINN} & \multicolumn{3}{c}{Probabilistic Benchmark} \\
        & Best & 2nd Best & 3rd Best & DeepAR & QRNN & NNQF \\
                \midrule
        Electricity & \textbf{0.1925}\tnote{1} & 0.1960\tnote{2} & 0.1974\tnote{3} & 0.2794 & 0.2020 & 0.2298\\
        Bike & \textbf{0.1856}\tnote{1} & 0.2228\tnote{3} & 0.2308\tnote{2} & 0.2317 & 0.3324 & 0.3808 \\
        OPSD & \textbf{0.0977}\tnote{3} & 0.1004\tnote{2} & 0.1159\tnote{1} & 0.1624 & 0.1554 & 0.1679 \\
        Price & \textbf{0.1833}\tnote{1} & 0.1837\tnote{4} & 0.1932\tnote{5} & 0.2023 & 0.2144 & 0.2576 \\
        \bottomrule
    \end{tabular}
    \begin{tablenotes}
        \item[1] \acrshort{cinn}-\acrshort{tft}
        \item[2] \acrshort{cinn}-\acrshort{nn}
        \item[3] \acrshort{cinn}-\acrshort{xgb}
        \item[4] \acrshort{cinn}-\acrshort{nhits}
        \item[5] \acrshort{cinn}-\acrshort{lr}
    \end{tablenotes}
    \end{threeparttable}
\end{table}

\paragraph{Quality} Concerning the quality of the probabilistic forecast, we select the three best-performing base forecasters in our approach and compare their performance to the performance of the selected probabilistic benchmark forecasting methods. \Cref{tab:crps_benchmark} reports the resulting mean \acrshort{crps} (\Cref{eq:crps}) on the test data set averaged over five runs for each data set. 

We observe that the best-performing base forecaster with a \acrshort{cinn} outperforms all probabilistic benchmarks for all data sets. Actually, in all four data sets, the three best-performing base forecasters combined with a \acrshort{cinn} outperform all benchmarks. 

\begin{table}[t]
    \centering
    \footnotesize
    \caption{The mean normalised width of the 98\%, 70\%, and 40\% \acrshortpl{pi} (\Cref{eq:pi_width}) for the cINN with the best-performing base forecaster from \Cref{subsec:selected_base} and the probabilistic benchmark forecasting methods from \Cref{subsec:probabilistc_benchmarks} over five runs. A low value indicates a sharp prediction interval.}
    \label{tab:width_benchmark}
    \begin{tabular}{ll|cccc}
        \toprule
        Prediction Interval & Data Set & \multicolumn{4}{c}{Prediction Interval Width} \\
        & & cINN & DeepAR & QRNN & NNQF \\
        \midrule
        & Electricity & 0.87 & 1.57 & 1.52 & 1.54 \\
        & Bike & 0.70 & 1.31 & 1.79 & 1.61 \\
        98\% & OPSD & 0.09 & 0.22 & 0.18 & 0.19 \\ 
        & Price & 0.50 & 0.53 & 0.86 & 0.83 \\
        \midrule
        & Electricity & 0.42 & 0.65 & 0.65 & 0.68 \\
        & Bike & 0.32 & 0.57 & 0.59 & 0.76 \\
        70\% & OPSD & 0.04 & 0.09 & 0.06 & 0.09 \\
        & Price & 0.22 & 0.23 & 0.30 & 0.37 \\
        \midrule
        & Electricity & 0.21 & 0.32 & 0.27 & 0.36 \\
        & Bike & 0.16 & 0.29 & 0.23 & 0.37 \\
        40\% & OPSD & 0.02 & 0.04 & 0.03 & 0.04 \\
        & Price  & 0.11 & 0.12 & 0.13 & 0.19 \\
        \bottomrule
    \end{tabular}
\end{table}

\paragraph{Sharpness} Regarding the sharpness of the probabilistic forecast, we compare normalised \acrshort{pi} width (\Cref{eq:pi_width}) of the best performing base forecaster combined with the \acrshort{cinn} and all probabilistic benchmark forecasting methods. \Cref{tab:width_benchmark} reports the mean normalised width of the 98\%, 70\%, and 40\% \acrshortpl{pi} on the test data set averaged over five runs. 

\begin{figure}
    \centering
    \scriptsize
    \input{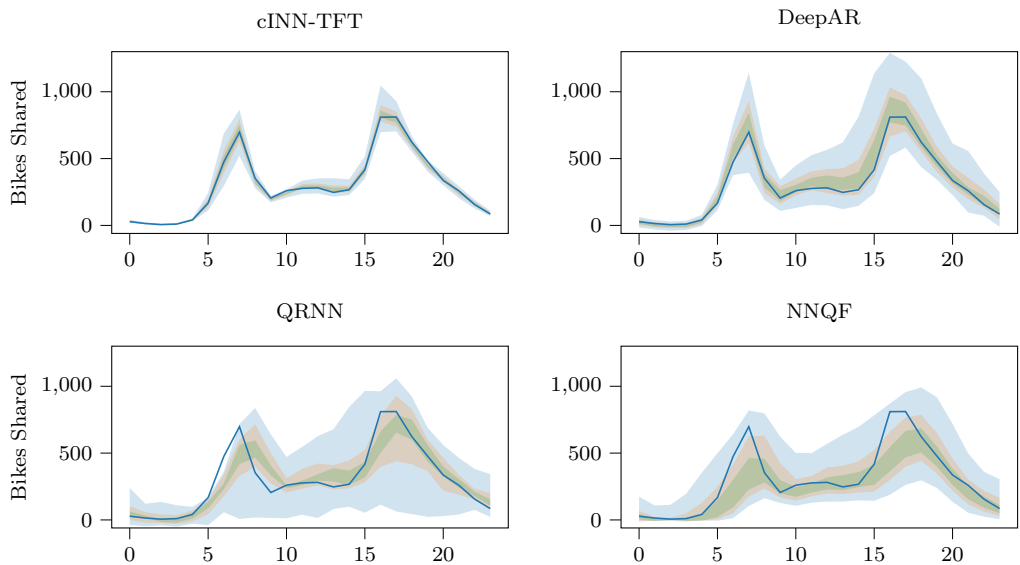}
    \caption{True values (blue), $98\%$ \acrshort{pi} (light blue), $70\%$ \acrshort{pi} (light yellow), and $40\%$ \acrshort{pi} (light green) for an exemplary $24$ hour forecast period on the test set of the bike data set.}
    \label{fig:pi_example}
\end{figure}

We observe that for all data sets and each of the three considered prediction intervals, the forecasts generated with the \acrshort{cinn} result in the narrowest \acrshortpl{pi}. This observation is confirmed by considering an exemplary graphical depiction of the \acrshortpl{pi} for a forecast period of $24$ hours from the bike data set in \Cref{fig:pi_example}. In this example, the \acrshortpl{pi} from the \acrshort{cinn} are narrower than those from all probabilistic benchmark forecasting methods and also accurately follow the true values.

\begin{figure}
    \centering
    \begin{tikzpicture}
    \begin{axis} [
        height = 5.5cm,
        width = 1\textwidth,
        ybar,
        ylabel=Training Time in Seconds,
        ylabel style={font=\footnotesize},
        table/col sep=comma,
        tick label style={font=\footnotesize},
        symbolic x coords ={cINN, DeepAR, QRNN, NNQF},
        xticklabel style={text width=1.25cm,align=center, font=\footnotesize},
        xtick=data,
        axis lines=left,
        enlarge x limits = 0.125,
        ymin=0,
        legend style = {
            font=\footnotesize,
            at={(0.45,1.4)},
            anchor=north,
        },
        legend cell align={left},
        legend columns=4
        ]
        \addplot+[error bars/.cd, y dir=both, y explicit]
            table[x=Method,y=Elec_mean, y error plus=Elec_std, y error minus=Elec_std] {results/Training_Times.csv};
        \addlegendentry{Electricity}
            
        \addplot+[error bars/.cd, y dir=both, y explicit]
            table[x=Method,y=Bike_mean, y error plus=Bike_std, y error minus=Bike_std] {results/Training_Times.csv};
        \addlegendentry{Bike}
        
        \addplot+[error bars/.cd, y dir=both, y explicit]
            table[x=Method,y=OPSD_mean, y error plus=OPSD_std, y error minus=OPSD_std] {results/Training_Times.csv};
        \addlegendentry{OPSD}
        
        \addplot+[error bars/.cd, y dir=both, y explicit]
            table[x=Method,y=Price_mean, y error plus=Price_std, y error minus=Price_std] {results/Training_Times.csv};
        \addlegendentry{Price}
        
    
    \end{axis}
\end{tikzpicture}
    \caption{The average training time across five training runs for the \acrshort{cinn} used to create probabilistic forecasts and all probabilistic benchmark forecasting methods from \Cref{subsec:probabilistc_benchmarks}. The error bars indicate the variance in the training time across the runs. Note that only the training time for the \acrshort{cinn} is recorded, the additional time required to train a deterministic base forecaster is not included.}
    \label{fig:training_time}
\end{figure}
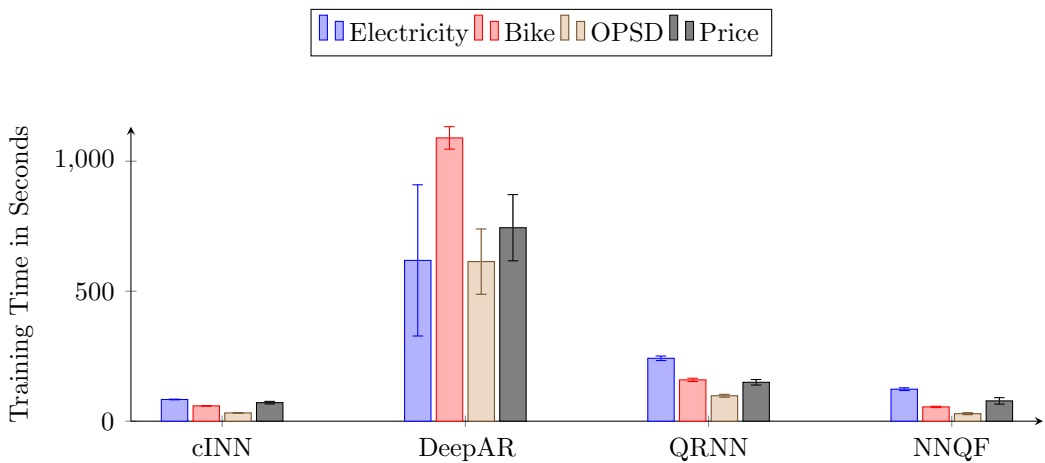

\paragraph{Training Time} Finally, we compare the training time of our approach and the probabilistic benchmark forecasting methods. In \Cref{fig:training_time}, we report the average training time across all five training runs and the variance across the runs. Importantly, we only report the training time for the \acrshort{cinn} and not the associated base forecaster. 

We observe that the \acrshort{cinn} requires significantly less training time than DeepAR or the \acrshort{qrnn}, and a similar time as the \acrshort{nnqf}.

\subsection{GEFCom2014 Probabilistic Price Forecasting}
\label{subsec:gefcom}
We lastly evaluate the proposed approach by retrospectively determining its placement in the \acrshort{gefcom}. The \acrshort{gefcom} was a probabilistic energy forecasting competition with four different forecasting tracks, namely load, price, wind, and solar forecasting \citep{hong2016probabilistic}. For the evaluation, we recreate the setup of the \acrshort{gefcom} price forecasting track in which 14 teams competed and compare the performance of our approach to the leading entrants from the competition. This comparison is based on the scoring mechanism from \acrshort{gefcom} that considers the final twelve of fifteen tasks, each creating 24-hour forecasts with 99\% quantiles. Given the pinball loss improvement $\operatorname{PL}_{\%}$ for each task, the final ranking is determined as the average relative pinball loss across all tasks weighted by the task number \citep{hong2016probabilistic}. For our approach, we apply all previous deterministic base learners and select the best-performing sampling parameter over the first three non-evaluated tasks and use this for all remaining tasks. The final weighted relative pinball loss and the resulting rank of our approach are shown in \Cref{tab:gefcom_small} (see \Cref{tab:gefcom_full} for the results per task).


\begin{table}[t]
    \centering
    \footnotesize
    \caption{The overall weighted pinball loss improvement all tasks and final rank in the GEFCom2014 price forecasting challenge for all base forecasters from \Cref{subsec:selected_base} combined with the \acrshort{cinn}. The weighted pinball loss improvement indicates the improvement of a given method over the GEFCom2014 baseline forecast weighted by the task number. The rank indicates the placement in GEFCom2014 according to the weighted relative pinball loss.}
    \label{tab:gefcom_small}
    \begin{threeparttable}
        \begin{tabular}{p{5cm}|c|c}
        \toprule
            & Average weighted pinball loss improvement & Rank  \\
            \midrule
            cINN-RF & 65.9 & \textbf{4}\tnote{1} \\
            cINN-LR & 65.7 & \textbf{4}\tnote{1} \\
            cINN-NN & 61.5 & \textbf{9}\tnote{1}\\
            cINN-XGBoost & 65.0 & \textbf{4}\tnote{1}\\
            cINN-N-HITS & 62.9 & \textbf{6}\tnote{1}\\
            cINN-TFT & 67.4 & \textbf{3}\\
            \midrule
            Tololo \citep{gaillard2016additive} & 71.7 & \textbf{1}\\
            Team Poland \citep{maciejowska2016hybrid} & 67.7 & \textbf{2} \\
            GMD \citep{dudek2016multilayer} & 67.1 & \textbf{3}\tnote{1} \\
            \bottomrule
        \end{tabular}
        \begin{tablenotes}
            \item[1] This ranking is determined by how each individual method would have placed in the GEFCom2014 price forecasting challenge in 2014 and, therefore, assumes that only one of the presented methods is considered at a time when creating the respective ranking.
        \end{tablenotes}
    \end{threeparttable}

\end{table}

Overall, our approach with a \acrshort{cinn} and various base forecasters performs well. With simple deterministic forecasting methods such as \acrshort{rf} regression and \acrshort{lr}, we achieve an average weighted pinball loss improvement that would have placed these methods within the top five of the competition. Furthermore, with a more advanced \acrshort{tft} base forecaster, we achieve an average weighted pinball loss improvement that would have resulted in a third-place finish. 

\section{Discussion}
\label{sec:discussion}
In this section, we first discuss our results and the implications these present, before we highlight some of the key insights gained from the evaluation.

\subsection{Results}
\label{subsec:discuss_results}
Regarding the results of our evaluation, we first discuss the forecasting performance, before considering the training time and the results in the \acrshort{gefcom}.


\paragraph{Forecasting Performance} Overall, our approach generates high-quality probabilistic forecasts. The approach not only outperforms a baseline that assumes Gaussian distributed errors, but also generally performs better in terms of \acrshort{crps} than three selected state-of-the-art probabilistic benchmark forecasting methods. Our approach also results in the narrowest \acrshortpl{pi}. However, despite being optimal according to \acrshort{crps}, the \acrshortpl{pi} created by our approach are quite narrow, which may not always be ideal. Therefore, it may be interesting to improve the calibration of the \acrshortpl{pi} generated by our approach.

\paragraph{Training Time} The training time of our approach is negligible in comparison to DeepAR and the \acrshort{qrnn}. However, we only report the training time of the used \acrshort{cinn} and not the associated base forecasters. In practice, we expect that an already existing trained deterministic base forecaster will be used. However, if this is not the case, the training time of the base forecaster must be taken into account, since these are highly dependent on the base forecaster used. Furthermore, the training time depends on the implementation used. Therefore, future work should consider how the reported training times vary if alternative libraries or programming languages are considered.



\paragraph{GEFCom2014} Not only does our approach perform competitively in the considered track of the \acrshort{gefcom}, but several factors undercut its true performance. All top-placing contestants in the competition perform specialised operations to improve forecasting performance, i.e. peak pre-processing \citep{gaillard2016additive}, filtering methods to weight certain days higher \citep{maciejowska2016hybrid}, or tailored training data periods to improve performance \citep{dudek2016multilayer}. In contrast, we consider all available data for training, refrain from complex pre-processing steps, and only use the default hyperparameters for the base forecasters. Furthermore, the true value of our approach is its ability to enable simple deterministic base forecasters such as a \acrshort{lr} or \acrshort{rf}, to rank within the top five compared to the original entrants. Finally, different base forecaster perform better for different tasks. Therefore, we expect an ensemble method that automatically selects the best base forecaster for a given task to deliver even better performance.

\subsection{Insights}
\label{subsec:discuss_insights}
In addition to the results, there are a few insights regarding the sampling parameter and flexible nature of our approach, which we discuss here.

\paragraph{Sampling Parameter} The sampling parameter $\sigma$ is currently manually selected to create optimal forecasts according to \acrshort{crps}. However, 
by varying this sampling parameter, it is possible to create different probabilistic forecasts which follow the same general shape but vary in the amount of uncertainty considered. Therefore, it may be interesting to investigate methods to automatically select an optimal sampling parameter given the observed data, a selected base forecaster, and a specific evaluation metric.

\paragraph{Flexible Nature} In the present article, we focus on evaluating the probabilistic forecasting performance of a single selected base forecaster combined with the \acrshort{cinn}. However, our proposed approach allows for a far more flexible application. Given a data set, we must only train the \acrshort{cinn} once before we apply it to generate probabilistic forecasts based on any arbitrary deterministic forecast. Furthermore, our proposed approach can be applied to create probabilistic forecasts without extending the existing deterministic forecasting method or applying extensive statistical knowledge. This is advantageous if a tailored deterministic forecasting method for a given scenario exists. In such a setting our approach will allow them to create probabilistic forecasts based on this tailored method without any further modification.


\section{Conclusion}
\label{sec:conclusion}
In the present article, we introduce an easy-to-use approach to create probabilistic forecasts from arbitrary deterministic forecasts by using a \acrshort{cinn} to learn the underlying distribution of the time series data. Our approach maps the underlying distribution of the data to a known and tractable distribution before combining the uncertainty from this known and tractable distribution with an arbitrary deterministic forecast to create probabilistic forecasts.

We show that our approach outperforms a probabilistic baseline forecast, which assumes that the errors of a deterministic forecast follow a Gaussian distribution. Moreover, we compare our approach to three state-of-the-art probabilistic benchmark forecasting methods, achieving the lowest \acrshort{crps} and probabilistic forecasts with the narrowest prediction intervals. Finally, we recreate the \acrshort{gefcom} and show that our approach enables simple deterministic base forecasters to rank within the top five.

Our approach offers a solution to generate flexible probabilistic forecasts based on arbitrary deterministic forecasts. In future work, this flexibility should be further investigated by automating the selection of the sampling parameter used to include uncertainty and considering how different metrics for optimising this parameter affect the resulting forecasts. Furthermore, it may be interesting to extend our approach to incorporate time series with stochastic inputs such as ensemble prediction systems from numerical weather prediction models.

\section*{Acknowledgements}
This project is funded by the Helmholtz Association’s Initiative and Networking Fund through Helmholtz AI and the Helmholtz Association under the Program “Energy System Design”.
\newpage
\appendix

\section{Implementation Details}

\subsection{Data Sets and Exogenous Features}

\begin{table}[htbp!]
    \centering
    \scriptsize
    \caption{Overview of the data sets used including the exogenous features considered, and the used train, validation, and test sets.}
    \label{tab:data_overview}
    \begin{threeparttable}
    \begin{tabular}{l|lllll}
        \toprule
        Data Set & Target & \makecell[l]{Exogenous \\ Features} & Train & Validation & Test \\
                \midrule
        Electricity & MT\_158 & \makecell[l]{Calendar\tnote{1}} &  $\left[0,18000\right)$ & $\left[18000,20000\right)$ &$\left[20000,22000\right]$  \\
        \\
        Bike\tnote{2} & cnt & \makecell[l]{Calendar\tnote{1},\\ Temperature, \\ Humidity, \\ Windspeed, \\ Weather Situation} & $\left[0,12000\right)$ & $\left[12000,14000\right)$ & $\left[14000,17544\right]$ \\
        \\
        OPSD & \makecell{load\_power \\ \_statistics} & \makecell[l]{Calendar\tnote{1}} & $\left[0,6000\right)$ & $\left[6000,7000\right)$ & $\left[7000,8400\right]$  \\
        \\
        Price & Zonal Price & \makecell[l]{Calendar\tnote{1}, \\ Forecast Total Load, \\ Forecast Zonal Load} &  $\left[0,13000\right)$ & $\left[13000,15000\right)$ & $\left[15000,21528\right]$ \\
        \bottomrule
    \end{tabular}
    \begin{tablenotes}
        \item[1] Sine- and cosine-encoded time of the day, the sine- and cosine-encoded month of the year, and a Boolean that indicates whether the current day is a weekend day or not.
        \item[2] To create a time index for this data, we merge the columns \emph{dteday} and \emph{hr} and deal with missing values using linear interpolation.
    \end{tablenotes}
    \end{threeparttable}
\end{table}
 \newpage
\subsection{Selected Base Forecasters}

\begin{table}[htbp!]
    \centering
    \footnotesize
    \caption{Overview of the selected base forecasters from \Cref{subsec:selected_base} used to create deterministic forecasts.}
    \label{tab:bf_overview}
    \begin{threeparttable}
            \begin{tabular}{l|lll}
        \toprule
        Base Forecaster & Classification & Implementation Details & Library Used \\
                \midrule
        \acrshort{rf} & Statistical & Default Hyperparameters &  SKlearn\tnote{1} \\
        \\
        \acrshort{lr} & Statistical & Default Hyperparameters & SKlearn\tnote{1} \\
        \\
        \acrshort{nn} & \makecell[l]{Machine \\ Learning} & \makecell[l]{\textbf{Hidden Layers:} 3\\ \textbf{Layer Sizes:} 90-64-32 \\ \textbf{Hidden Activation} \\ \textbf{Function:} relu \\ \textbf{Output Activation} \\ \textbf{Function:} linear \\ \textbf{Optimiser:} Adam\tnote{2} \\ \textbf{Batch Size:} 100 \\ \textbf{Max Epochs:} 100} & \makecell[l]{Tensorflow\tnote{3}  \\ Keras\tnote{4}}  \\
        \\
        \acrshort{xgb} & \makecell[l]{Gradient \\ Boosting} & Default Hyperparameters &  XGBoost\tnote{5} \\
        \\
        \acrshort{nhits} & Deep Learning & Default Hyperparameters & PyTorch Forecasting\tnote{6} \\
        \\
        \acrshort{tft} & Deep Learning & Default Hyperparameters & PyTorch Forecasting\tnote{6} \\
        \bottomrule
    \end{tabular}
    \begin{tablenotes}
    \scriptsize
        \item[1] \citep{Pedregosa2011}
        \item[2] \citep{Kingma2015}
        \item[3] \url{https://www.tensorflow.org/}
        \item[4] \url{https://keras.io/}
        \item[5]  \citep{Chen2016}
        \item[6] \url{https://pytorch-forecasting.readthedocs.io/en/stable/index.html}
    \end{tablenotes}
    \end{threeparttable}
\end{table}

\newpage
\subsection{Selected Sampling Parameters for Evaluation}

\begin{table}[htbp!]
    \centering
    \footnotesize
    \caption{The selected sampling parameter for each base forecaster from \Cref{subsec:selected_base} and each data set used in the evaluation.}
    \label{tab:sampling_main}
    \begin{tabular}{l|cccccc}
        \toprule
        Data Set & RF & LR & NN & XGBoost & N-HITS & TFT \\
                \midrule
        Electricity & 0.600 & 0.500 & 0.500 & 0.500 & 0.600 & 0.600 \\
        Bike & 0.800 & 0.350 & 0.350 & 0.300 & 0.300 & 0.350 \\
        OPSD & 0.300 & 0.350 & 0.250 & 0.1250 & 0.250 & 0.300 \\
        Price & 0.750 & 0.750 & 0.700 & 0.450 & 0.850 & 0.850 \\
        GEFCom Competition &  0.950 & 1.050 & 0.900 & 0.500 & 0.750 & 1.050 \\
        \bottomrule
    \end{tabular}
\end{table}

\newpage
\subsection{Implementation details for the used cINN}\label{appendix:implementation_cinn}
When training the use \acrshort{cinn} we use the Adam optimiser with a maximum of 100 Epochs. Furthermore, when sampling in the latent space to create probabilistic forecasts we consider a sample size of 100.

\begin{table}[htbp!]
\footnotesize
    \caption{The architecture of the used \acrshort{cinn}}
    \label{tab:cINNHyperparams}
    \centering
    \begin{tabular}{ll}
    \toprule
        Parameter & Description \\
        \midrule
         Layers per block & Glow coupling layer and random permutation  \\
         Subnetwork in block & Fully connected, see \Cref{tab:implementation_cINN} \\
         Number of blocks & 5 \\
         Conditional network & Fully connected, see \Cref{tab:implementation_cINN} \\
        \bottomrule
    \end{tabular}
\end{table}

\begin{table}[htbp!]
\footnotesize
    \caption{Implementation details of the subnetwork and the conditioning network $q$ in the used cINN.}
    \label{tab:implementation_cINN}
    \begin{subtable}{\textwidth}
        \caption{Subnetwork}
        \label{tab:implementation_cINN_sub}
        \centering
        \begin{tabular}{ll}
            \toprule
            Layer & Description\\
            \midrule
            Input & [Output of previous coupling layer, \\
                  & conditional information] \\
            1 &  Dense 32 neurons; activation: tanh\\
            2 &  Dense 24 neurons; activation: linear\\
            \bottomrule
        \end{tabular}    
    \end{subtable}
    \par\bigskip
    \begin{subtable}{\textwidth}
        \caption{Conditioning network}
        \label{tab:implementation_cINN_cond}
        \centering
        \begin{tabular}{ll}
            \toprule
            Layer & Description\\
            \midrule
            Input & [Calendar information, \\
                  & historical information, \\
                  & exogenous forecasts if available] \\
            1 &  Dense 8 neurons; activation: tanh\\
            2 &  Dense 4 neurons; activation: linear\\
            \bottomrule
        \end{tabular}    
    \end{subtable} 
\end{table}

\newpage
\section{Improvement Over Baseline}

\begin{table}[htbp!]
    \centering
    \footnotesize
    \caption{Comparison of the absolute \acrshort{crps} between the baseline assuming a Gaussian distribution and the same base forecaster from \Cref{subsec:selected_base} with the cINN. We report the mean \acrshort{crps} averaged across five runs for the test data on all data sets.}
    \label{tab:crps_raw}
    \begin{tabular}{ll|cccc}
    \toprule
     Base Forecaster & Variant & Electricity & Bike & OPSD & Price \\
     \midrule
    \multirow{2}{*}{\acrshort{rf}}   & Baseline & 0.2128 & 0.2584 & 0.1259 & 0.2410 \\ 
     & cINN & 0.1976 & 0.2322 & 0.1252 & 0.2223 \\ 
     \midrule
    \multirow{2}{*}{\acrshort{lr}}   & Baseline & 0.3487 & 0.3595 & 0.1853 & 0.3047 \\ 
     & cINN & 0.3020 & 0.3265 & 0.1607 & 0.1932 \\ 
     \midrule
    \multirow{2}{*}{\acrshort{nn}}   & Baseline & 0.2496 & 0.2429 & 0.1034 & 0.2305 \\ 
     & cINN & 0.1960 & 0.2308 & 0.1004 & 0.1954 \\
     \midrule
    \multirow{2}{*}{XGBoost}   & Baseline & 0.2142 & 0.2390 & 0.0995 & 0.2240 \\ 
     & cINN & 0.1960 & 0.2228 & 0.0977 & 0.2000\\ 
     \midrule
    \multirow{2}{*}{N-HITS} & Baseline & 0.2950 & 0.2638 & 0.1434 & 0.2615 \\
     & cINN & 0.2099 & 0.2534 & 0.1319 & 0.1837 \\ 
     \midrule
    \multirow{2}{*}{\acrshort{tft}}   & Baseline & 0.2693 & 0.2150 & 0.1257 & 0.2389\\ 
     & cINN & 0.1925 & 0.1856 & 0.1159 & 0.1833 \\ 
     \bottomrule
    \end{tabular}
\end{table}


\begin{sidewaystable}[htbp!]
\section{Full GEFCom2014 Results}
    \centering
    \scriptsize
    \caption{GEFCom2014 results for each task. The performance of each base forecaster from \Cref{subsec:selected_base} combined with the \acrshort{cinn} is reported, as well as the performance of the best three entrants from 2014. For each task, pinball loss improvement $\operatorname{PL}_{\%}$ compared to the GEFCom2014 baseline is calculated. The final results are a linear weighted average of the pinball loss improvement in each task, weighted by the task number.}
    \label{tab:gefcom_full}
        \begin{threeparttable}
        \begin{tabular}{l|cc|cc|cc|cc|cc|cc|cc}
                Model & \multicolumn{2}{c}{Task 1} & \multicolumn{2}{c}{Task 2}  & \multicolumn{2}{c}{Task 3}  & \multicolumn{2}{c}{Task 4}  & \multicolumn{2}{c}{Task 5}  & \multicolumn{2}{c}{Task 6} \\
                & PL & $\operatorname{PL}_{\%}$ & PL & $\operatorname{PL}_{\%}$ & PL & $\operatorname{PL}_{\%}$ & PL & $\operatorname{PL}_{\%}$ & PL & $\operatorname{PL}_{\%}$ & PL & $\operatorname{PL}_{\%}$ \\
                \midrule
                cINN-RF & 1.10 & 72.67 & 3.46 & 56.61 & 2.93 & 48.64 & 7.13 & 41.35 & 5.24 & 86.34 & 6.45 & 85.41 \\ 
                 cINN-LR & 1.76 & 56.25 & 2.95 & 62.97 & 1.92 & 66.36 & 2.97 & 75.53 & 8.59 & 77.60 & 12.14 & 72.55 \\
                 cINN-NN & 1.16 & 71.14 & 4.38 & 45.01 & 2.02 & 64.57 & 4.31 & 64.53 & 7.26 & 81.05 & 7.10 & 83.96 \\
                 cINN-XGBoost & 1.70 & 57.69 & 3.53 & 55.76 & 2.74 & 51.90 & 5.21 & 57.12 & 6.08 & 84.14 & 7.65 & 82.71 \\
                 cINN-N-HITS & 1.70 & 57.87 & 2.45 & 69.21 & 1.05 & 81.57 & 3.75 & 69.15 & 8.03 & 79.06 & 10.08 & 77.21 \\
                 cINN-TFT & 1.22 & 69.80 & 2.50 & 68.64 & 1.16 & 79.67 & 3.08 & 74.65 & 6.30 & 83.57 & 9.20 & 79.21 \\
                \midrule
                Tololo\tnote{2} & 1.71 & 57.62 & 1.45 & 81.80 & 1.10 & 80.65 & 2.02 & 83.40 & 9.16 & 76.11 &	4.68 & 89.41 \\
                Team Poland\tnote{3} & 1.97 & 50.98 & 1.82 & 77.19 &	1.19 & 79.11 & 2.82 & 76.77 & 7.56 & 80.28 & 4.21 & 90.49 \\
                GMD\tnote{4} & 3.73 & 7.48 & 1.78 & 77.63 & 0.92 & 83.84 & 5.09 & 58.12 & 6.21 & 83.79 & 3.83 & 91.35 \\
                \bottomrule 
                \\
                \\
                \toprule
                 & \multicolumn{2}{c}{Task 7} & \multicolumn{2}{c}{Task 8}  & \multicolumn{2}{c}{Task 9}  & \multicolumn{2}{c}{Task 10}  & \multicolumn{2}{c}{Task 11}  & \multicolumn{2}{c}{Task 12}   & \multicolumn{2}{c}{Overall}  \\
                 & PL & $\operatorname{PL}_{\%}$ & PL & $\operatorname{PL}_{\%}$ & PL & $\operatorname{PL}_{\%}$ & PL & $\operatorname{PL}_{\%}$ & PL & $\operatorname{PL}_{\%}$ & PL & $\operatorname{PL}_{\%}$  & Total Weighted $\operatorname{PL}_{\%}$ & Rank \\
                \midrule
                 cINN-RF & 7.55 & 58.54 & 2.51 & 92.06 & 2.16 & 94.97 & 1.95 & 31.89 & 1.79 & 44.03 & 5.69 & 74.56 & \emph{65.9} & \textbf{4}\tnote{1}\\
                 cINN-LR & 12.66 & 30.53 & 1.85 & 94.14 & 2.39 & 94.44 & 1.50 & 47.36 & 1.66 & 48.33 & 7.54 & 66.32  & \emph{65.7} & \textbf{4}\tnote{1}\\
                 cINN-NN & 6.69 & 63.29 & 2.83 & 91.03 & 1.62 & 96.21 & 1.88 & 34.33 & 1.70 & 46.99 & 15.38 & 31.30  & \emph{61.5} & \textbf{9}\tnote{1}\\
                 cINN-XGBoost & 7.64 & 58.05 & 2.43 & 92.30 & 1.47 & 96.58 & 2.03 & 28.87 & 1.83 & 42.97 & 7.05 & 68.50  & \emph{65.0} & \textbf{4}\tnote{1}\\
                 cINN-N-HITS & 11.70 & 35.79 & 1.40 & 95.58 & 1.28 & 97.01 & 2.31 & 19.17 & 2.06 & 35.72 & 6.37 & 71.53 &  \emph{62.9} & \textbf{6}\tnote{1}\\
                 cINN-TFT & 8.53 & 53.21 & 1.67 & 94.71 & 1.43 & 96.67 & 2.42 & 15.41 & 1.66 & 48.18 & 4.95 & 77.88 & \emph{67.4} & \textbf{3}\\
                \midrule
                Tololo\tnote{2} & 1.60 & 91.24 & 0.75 & 97.61 & 2.46 & 94.27 & 2.96 & -3.70 & 1.35 & 57.98 & 3.56 & 84.10 & \emph{71.7} & \textbf{1}\\
                Team Poland\tnote{3} & 2.60 & 85.75 & 1.05 & 96.68 & 1.24 & 97.11 & 4.06 & -42.17 & 1.08 & 66.15 & 3.07 & 86.31 & \emph{67.7} & \textbf{2} \\
                GMD\tnote{4} & 4.93 & 72.92 & 1.48 & 95.32 & 1.66 & 96.14 & 2.06 & 27.82 & 2.12 & 33.72 & 6.85 & 69.41 & \emph{67.1} & \textbf{3}\tnote{1} \\
                \bottomrule 
        \end{tabular}
        \begin{tablenotes}
            \item[1] This ranking is determined by how each individual model would have placed in the GEFCom2014 price forecasting challenge in 2014 and therefore assumes that the other models introduced in this article are not included.
            \item[2] \citet{gaillard2016additive}
            \item[3] \citet{maciejowska2016hybrid}
            \item[4] \citet{dudek2016multilayer}
        \end{tablenotes}
        \end{threeparttable}    
\end{sidewaystable}

\newpage

\bibliographystyle{elsarticle-harv} 
\bibliography{bibliography.bib}

\end{sloppypar}
\end{document}